\documentclass{article}
\usepackage[utf8]{inputenc}
\usepackage{authblk}
\usepackage{setspace}
\usepackage[margin=1.25in]{geometry}
\usepackage{graphicx}
\graphicspath{ {./figures/} }
\usepackage{subcaption}
\usepackage{amsmath}
\usepackage{xspace}
\usepackage{amssymb,amsfonts}
\usepackage{textcomp}
\usepackage{xcolor}
\usepackage{colortbl}
\usepackage{multirow}
\usepackage{booktabs}
\usepackage{utfsym}
\usepackage{makecell}
\usepackage{float}
\usepackage{afterpage}
\usepackage{lineno}
\newcommand{\ourapproach}{\textsc{S$^2$ALM}\xspace}

\definecolor{ForestGreen}{RGB}{73,200,94}

\usepackage[style=nejm, 
citestyle=numeric-comp,
sorting=none]{biblatex}
\title{\ourapproach: Sequence-Structure Pre-trained Large Language Model for Comprehensive Antibody Representation Learning}

\author[1,2$\dag$]{Mingze Yin}
\author[1$\dag$]{Hanjing Zhou}
\author[3]{Jialu Wu}
\author[1]{Yiheng Zhu}
\author[2]{Yuxuan Zhan}
\author[1]{Zitai Kong}
\author[3]{Hongxia Xu}
\author[3$\ast$]{Chang-Yu Hsieh}
\author[4$\ast$]{Jintai Chen}
\author[3$\ast$]{Tingjun Hou}
\author[5,6,7$\ast$]{Jian Wu}
\affil[1]{College of Computer Science and Technology, Zhejiang University, Hangzhou, China.}
\affil[2]{School of Medicine, Zhejiang University, Hangzhou, China.}
\affil[3]{Innovation Institute for Artificial Intelligence in Medicine of Zhejiang University, College of Pharmaceutical Sciences, Zhejiang University, Hangzhou, China.}
\affil[4]{Department of Computer Science, UIUC, Illinois, USA}
\affil[5]{State Key Laboratory of Transvascular Implantation Devices of The Second Affiliated Hospital, Zhejiang University School of Medicine, Hangzhou, China.}
\affil[6]{School of Public Health, Zhejiang University, Hangzhou, China.}
\affil[7]{Zhejiang Key Laboratory of Medical Imaging Artificial Intelligence, Hangzhou, China.}
\affil[$\dag$]{These authors contributed equally to this work.}
\affil[*]{Correspondence to: 
tingjunhou@zju.edu.cn (T. Hou); 
kimhsieh@zju.edu.cn (C. Hsieh);
jtchen721@gmail.com (J. Chen); 
wujian2000@zju.edu.cn (J. Wu)}
\date{}

\begin{document}

\maketitle

%%%%%% Abstract %%%%%%
\begin{abstract}
Antibodies safeguard our health through their precise and potent binding to specific antigens, demonstrating promising therapeutic efficacy in the treatment of numerous diseases, including COVID-19. Recent advancements in biomedical language models have shown the great potential to interpret complex biological structures and functions.
However, existing antibody specific models have a notable limitation that they lack explicit consideration for antibody structural information, despite the fact that both 1D sequence and 3D structure carry unique and complementary insights into antibody behavior and functionality.
This paper proposes \textbf{S}equence-\textbf{S}tructure multi-level pre-trained \textbf{A}ntibody \textbf{L}anguage \textbf{M}odel (\ourapproach), combining holistic sequential and structural information in one unified, generic antibody foundation model. 
We construct a hierarchical pre-training paradigm incorporated with two customized multi-level training objectives to facilitate the modeling of comprehensive antibody representations. \ourapproach's representation space uncovers inherent functional binding mechanisms, biological evolution properties and structural interaction patterns. 
Pre-trained over 75 million sequences and 11.7 million structures, \ourapproach can be adopted for diverse downstream tasks: accurately predicting antigen-antibody binding affinities, precisely distinguishing B cell maturation stages, identifying antibody crucial binding positions, and specifically designing novel coronavirus-binding antibodies. 
Remarkably, \ourapproach outperforms well-established and renowned baselines and sets new state-of-the-art performance across extensive antibody specific understanding and generation tasks. 
\ourapproach's ability to model comprehensive and generalized representations further positions its potential to advance real-world therapeutic antibody development, potentially addressing unmet academic, industrial, and clinical needs.
\end{abstract}

%%%%%% Main Text %%%%%%
\section{Introduction}
% 抗体介绍
% An antibody is a Y-shaped molecule composed of two identical heavy chains and two identical light chains linked by disulfide bonds. For each chain, there exist a variable domain and several constant domains. 
Antibodies, also known as immunoglobulins, are specialized proteins produced by the immune system to protect against a range of diseases (\textit{e.g.}, SARS-CoV-2), which fulfill the responsibility as guardians in the human body via evolving to complement their structures with the corresponding antigen structures~\cite{antibody_1, antibody_2, antibody_3}. 
% 抗体药物的重要性
Due to the high specificity and low adverse effect of antibodies, antibody drugs have important clinical significance and account for nearly one-fifth of new drug approvals by the FDA each year~\cite{FDA}.
% clinical relevance
By mimicking the actions of the immune system, these drugs specifically target harmful agents like viruses and cancer cells, to detect viral infections or stimulate T-cell immunity in cancer treatment~\cite{IgLM, PALM, KEDD, BALM}.
Therefore, deciphering the information stored in antibody sequences and structures is crucial for understanding immune responses, disease development, and may accelerate the development of therapeutic antibodies for broad-spectrum disease treatment.

% 为什么需要计算建模的方法？
To alleviate the burden of time-consuming wet-lab experiments, in recent years, the computer-aided antibody engineering has emerged to improve the efficiency of antibody evolution and identify promising therapeutic antibodies with desirable developability profiles.
% traditional approaches
While interpretable, traditional representation learning approaches depend on inefficient hand-crafted features that may miss hidden or latent patterns in diverse data. 
% 引出抗体特异性大语言模型
Drawing inspiration from Natural Language Processing (NLP) with the introduction of pre-trained foundational models, similar AI technologies have emerged as the pivotal technology to interpret the language of biology~\cite{Evo, EnsemPPIS, DeepSecE, MIPPI}. 
Pre-trained on large-scale antibody corpora, Antibody specific large Language Models (ALMs)~\cite{AntiBERTa, AntiBERTy, AbLang, EATLM, PALM, IgLM} have exhibited powerful capabilities in advancing the understanding of antibody structures and functions. Additionally, ALMs consistently outperform general-purpose Protein Language Models (PLMs) because the mechanism of antibody evolution is strongly biased to the target antigen. This finding inspires us to dive deep into the development of comprehensive antibody specific pre-training.

\begin{figure}[!ht]
    \centering
    \includegraphics[width=0.98\linewidth]{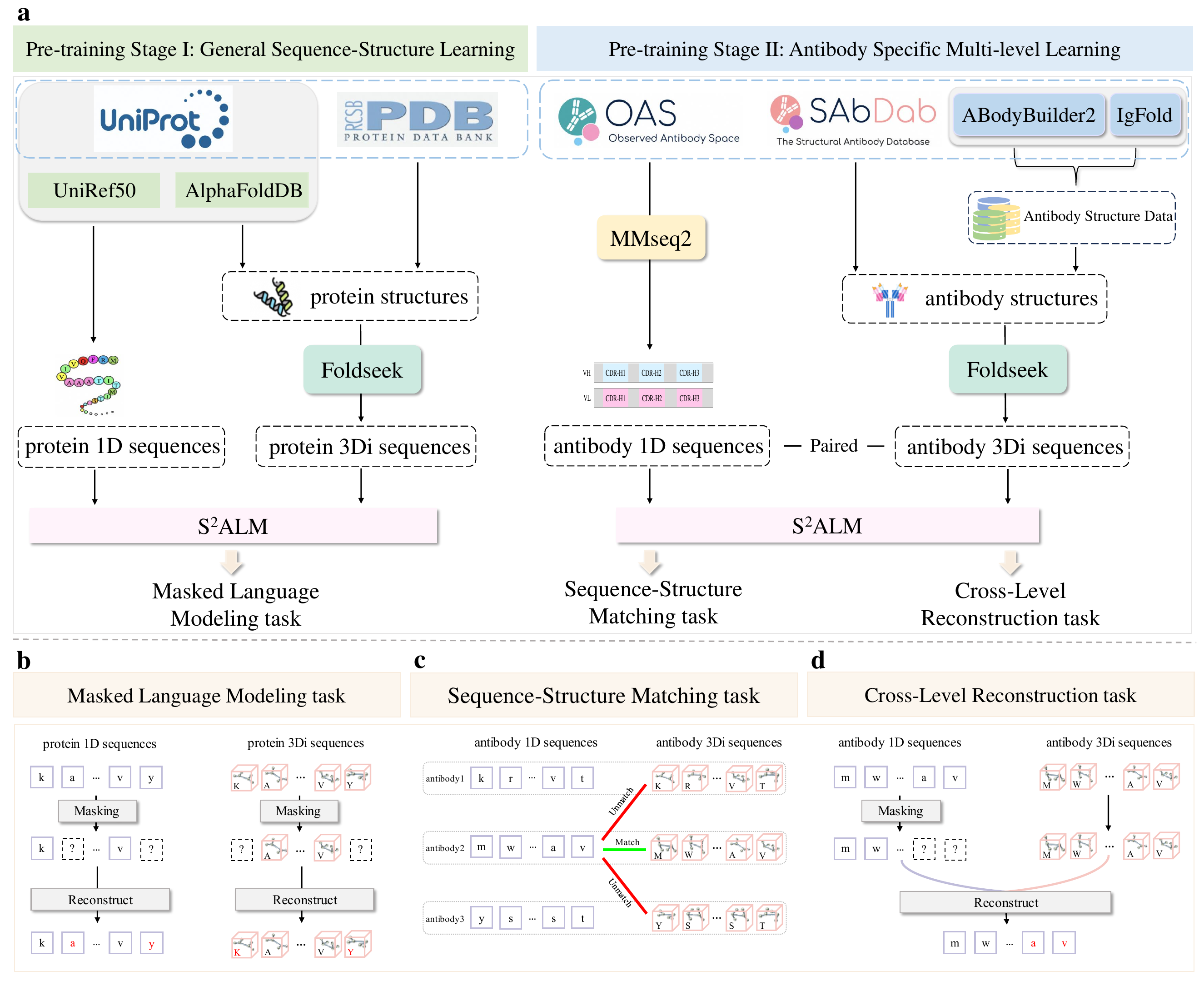}
    \caption{\textbf{Overview of the proposed hierarchical pre-training paradigm containing two stages.} \textbf{a}, In stage I, \ourapproach aims at general sequence-structure learning with protein sequences and structures. In stage II, \ourapproach learns antibody specific multi-level knowledge using antibody sequences and structures. \textbf{b}, Masked Language Modeling (MLM) reconstructs the masked tokens based on the contextualized information. \textbf{c}, Sequence-Structure Matching (SSM) identifies the matching relationships between 1D and 3Di sequences. \textbf{d}, Cross-Level Reconstruction (CLR) reconstructs the corrupted tokens based on hybrid information from both 1D and 3Di sequences.}
    \label{Fig:framework}
\end{figure}

% 结构信息的重要性
The central tenet of molecular biology is that an antibody's amino acid sequence determines its three-dimensional structure, and the three-dimensional structure determines its biological function. The spatial structure plays an integral role for antibody functional characterization, as the principle of antigen-antibody structural binding indicates that antibody structures directly determine biological functions. Consequently, injecting structural information into antibody pre-training emerges a highly compelling endeavor.
In this paper, we propose \textbf{S}equence-\textbf{S}tructure multi-level pre-trained \textbf{A}ntibody \textbf{L}anguage \textbf{M}odel (\ourapproach), incorporating antibody sequential and structural information to construct a comprehensive antibody foundation model. 
To accomplish this, we collect a holistic pre-training dataset with massive 1D sequences and 3D structures, and propose a hierarchical pre-training paradigm (Fig.~\ref{Fig:framework}), accompanied by two customized training objectives: Sequence-Structure Matching (SSM) and Cross-Level Reconstruction (CLR). 
These methodologies allow the model to concurrently process and analyze antibody sequence and structure data, further modeling comprehensive antibody representations to facilitate the deciphering of the biological language.

Through extensive interpretability analyses, \ourapproach exhibited an emergent understanding of unobserved patterns inherent in antibody structures and functions. Furthermore, to illustrate \ourapproach's practical effectiveness, we conducted in-depth examinations to assess the performance of \ourapproach across a range of downstream scenarios for antibody analysis. The outstanding performance verifies the \ourapproach's substantial potential to serve as the novel antibody foundation model and highlights its applicability in real-world therapeutic development.

% contributions
The main contributions of \ourapproach are three-fold:
\begin{itemize}
\item[$\bullet$]
Pre-trained on a large-scale multi-level dataset containing 75 million 1D sequences and 11.7 million 3D structures from protein and antibody domains, \ourapproach learns comprehensive antibody representations uncovering inherent patterns within structures and functions.

\item[$\bullet$]
A hierarchical pre-training paradigm incorporated with two customized training objectives, Sequence-Structure Matching (SSM) and Cross-Level Reconstruction (CLR), is proposed to promote the injection of antibody structural information.

% therapeutic antibody development and immune response analysis
\item[$\bullet$]
\ourapproach consistently exceeds state-of-the-art performance on extensive tasks including antigen binding prediction, B cell maturation analysis, antibody paratope prediction, binding affinity prediction and computational antibody design, exhibiting its broad applicability in therapeutic antibody development and immune response analysis.
\end{itemize}

\section{Materials and Methods}
% experimentally-determined structures
% computationally-predicted structures
\subsection{Pre-training Data}
\begin{figure}[htbp]
    \centering
    \includegraphics[width=0.98\linewidth]{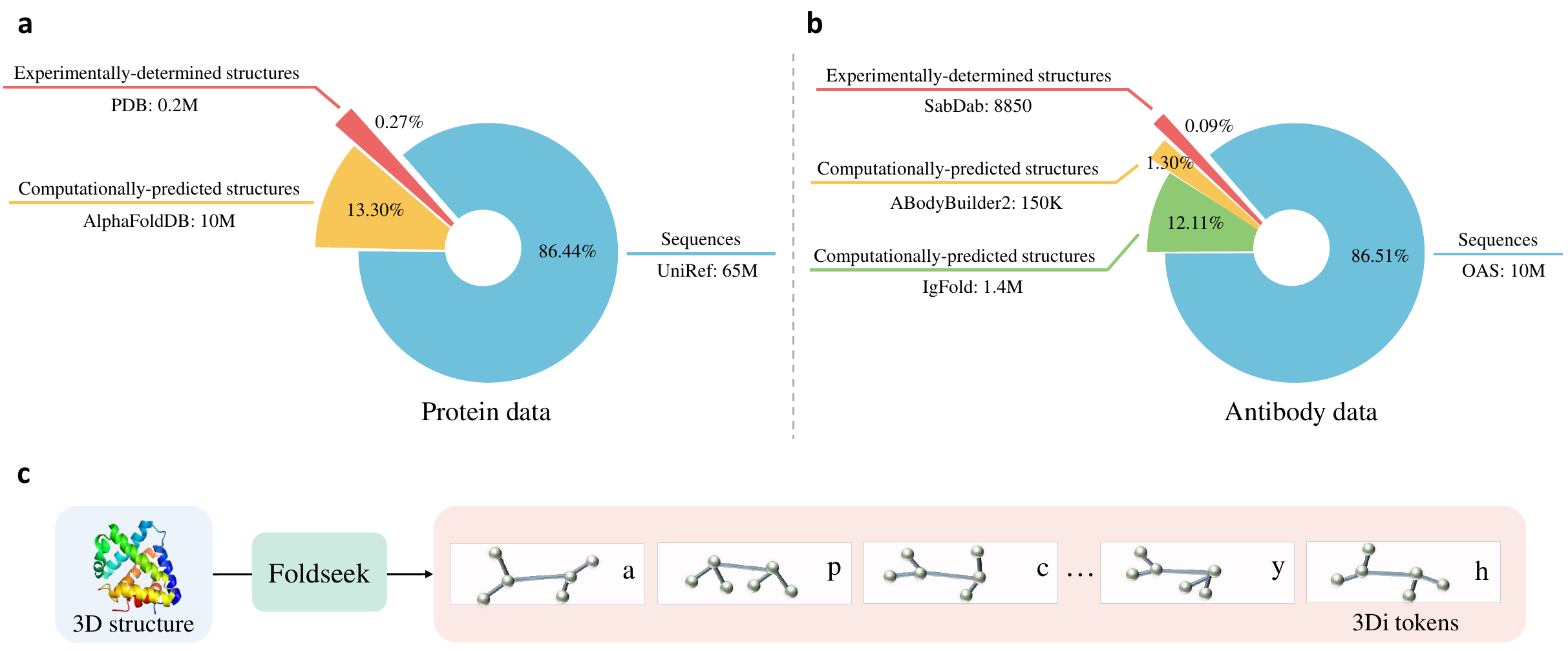}
    \caption{\textbf{Illustrations of compositional ratios of the pre-training data and the structural encoding protocol.}
    \textbf{a},
    The protein data contains three parts: sequences, experimentally-determined structures, computationally-predicted structures. 
    \textbf{b},
    The antibody data contains four parts: sequences, experimentally-determined structures, computationally-predicted structures from ABodyBuilder2~\cite{ABodyBuilder2} and IgFold~\cite{IgFold}.
    \textbf{c}, Efficient encoding protocol of protein 3D structures. Foldseek~\cite{foldseek} is employed to discretize the target 3D structure into the 3Di sequences, deciphering 3Di states which describe the tertiary interaction between a residue and its nearest neighbor.}
    \label{Fig:data_pie}
\end{figure}

% encompass various levels and span multiple domains
In the field of antibody pre-training, there is an abundance of sequential data, whereas the quantity of structure data remains limited, especially those determined by experiments. To compensate for the inadequacy of experimentally-determined antibody structures, we additionally introduce computationally-predicted antibody structures and general protein structures for comprehensive large-scale pre-training. Eventually, our holistic pre-training data encompasses various levels and spans multiple domains, including 75 million 1D sequences and 11.7 million 3D structures from protein and antibody domains. Fig.~\ref{Fig:data_pie}a-b presents compositional ratios of various parts in the pre-training dataset, and details of each part are as follows.

\subsubsection{Protein Sequence and Structure Data}
% UniRef: 65 million
% PDB: 0.2 million
% AlphaFoldDB: 10 million
% filter
Extensive protein sequence and structure data is incorporated for general sequence-structure learning.
Specifically, for the primary structural information (\textit{i.e.}, sequential information), 65 million protein sequences are derived from UniRef50~\cite{Uniref}, which is a clustering of UniRef90 seed sequences at 50\% sequence identity. Moreover, to alleviate the insufficiency of antibody structure data, the extra protein structure data is incorporated. For the secondary and tertiary structural information, we utilize 0.2 million experimentally-determined general protein 3D structures sourced from Protein Data Bank (PDB)~\cite{PDB} and randomly sample 10 million computationally-predicted protein 3D structures from AlphaFold Protein Structure Database (AFDB)~\cite{AFDB}.

\subsubsection{Antibody Sequence Data}
% OAS: 10 million
The pre-training antibody sequence data comes from the Observed Antibody Space (OAS) database~\cite{OAS}, which contains over two billion unpaired sequences and two million paired sequences of antibody heavy and light chains. The OAS database boasts a rich collection of antibody sequences from 825 unique subjects spanning across six different species, where humans account for 88\% and mice account for 11\%. We downloaded the OAS database of unpaired and paired antibody sequences on January 11th, 2024, and pre-processed the antibody sequence data by filtering, cleaning and clustering. First, all duplicate antibody sequences were filtered out and removed. Additionally, we excluded antibody sequences containing unusual residues (Selenocysteine, Pyrrolysine), along with those having any framework region shorter than IMGT defined. Furthermore, to mitigate data leakage, we clustered antibody sequences at 70\% sequence identity over the whole sequence using the MMseq2 algorithm~\cite{MMseq2}. Alignment coverage in MMseq2 was calculated with respect to the target sequence (``cov-mode 1''), with all other parameters setting to their default values. Ultimately, the antibody sequence data comprises 216,437,989 unique antibody sequences, from which we randomly select 10 million sequences for training.

\subsubsection{Antibody Structure Data}
% SabDab: 8850 (10k)
% ABodyBuilderDB: 150k
% IgFoldDB: 1.4 million
% Since current ALMs lack explicit consideration for antibody structural information, we innovatively propose to integrate antibody specific structural knowledge to achieve the multi-level pre-training.
Given the central tenet that antibody structure significantly governs its function, we innovatively explore to inject antibody specific structural knowledge into ALMs. To achieve this, we collect experimentally-determined antibody structure data based on Structural Antibody Database (SabDab)~\cite{SabDab}, which is composed of approximately 10 thousand antibody structures retrieved from PDB~\cite{PDB}. Considering the limited quantity of experimentally-determined antibody structures, we seek antibody structure prediction models for assistance to provide ample computationally-predicted structures. Concretely, 150 thousand antibody structures predicted by ABodyBuilder2~\cite{ABodyBuilder2} and 1.4 million antibody structures predicted by IgFold~\cite{IgFold} are incorporated for pre-training. Overall, the aforementioned three data sources collectively contribute to antibody structure learning.

% efficient structure encoding technique
\subsection{Structural Encoding Technique}
Foldseek~\cite{foldseek} is a computational tool for fast and accurate protein structure search, merging features such as amino acid spatial angles, distances, and sequence positions. It utilizes the VQ-VAE model~\cite{VQ-VAE} to encode the 3D protein structure as distinct and information-rich 3D interaction (3Di) tokens. Each amino acid is assigned a token based on its distance and relative position to the nearest amino acids in the folded protein structure. Foldseek achieves this transformation through the identification of the nearest neighbors and the extraction of distinctive features for individual residues~\cite{SaProt}.
% 3Di tokens, 3Di sequences
In this paper, we introduce Foldseek to accomplish the efficient encoding of 3D structures, as depicted in Fig.~\ref{Fig:data_pie}c. 
% facilitating sequence-structure large-scale pre-training, and promoting comprehensive antibody representation learning. 
Concretely, we adopt Foldseek with the default setting of 20 3Di tokens. Aligning all residue sites, protein and antibody 3D structures are transformed into 3Di sequences (\textit{i.e.}, 1D sequences storing 3D structural information). Such technique fills the void in effective encoding of structure data and enables \ourapproach to handle sequence-structure multi-level information in a hybrid and unified manner, promoting comprehensive antibody representation learning.
% This solves the challenge of efficiently encoding 3D structures, enabling the integration of structural information into the large-scale antibody pre-training process.

\subsection{Hierarchical Pre-training Paradigm}
\subsubsection{Stage I: General Sequence-Structure Learning}
\label{sec:stage1}
% 1、过程：multi-level vocabulary + MLM_loss
% 2、强调一下为什么要加protein数据，第一阶段目的
In pre-training stage I, we first tokenize the sequences and structures of proteins by an innovative multi-level vocabulary $\mathcal{F} = \mathcal{V}_1 + \mathcal{V}_2$. 
The original sequence alphabet $\mathcal{V}_1$ consists of 20 standard amino acids. A protein or antibody sequence can be denoted as $s=\{s_1, s_2,\dots,s_n\}$, where $s_i \in \mathcal{V}_1$ represents the 1D residue token at the $i_{th}$ site.
Building upon the concept of Foldseek~\cite{foldseek}, the newly-built structure alphabet $\mathcal{V}_2$ utilizes 20 distinct 3Di tokens to accomplish the generation of pseudo structural sequences (\textit{i.e.}, 3Di sequences). A protein or antibody structure can be denoted as $a=\{a_1,a_2,\dots,a_m\}$, with $a_j \in \mathcal{V}_2$ indicating the 3Di token at the $j_{th}$ site.

Building on the multi-level vocabulary, we obtain 1D and 3Di sequences and feed them into the model alternately. 
During pre-training stage I, we train \ourapproach using the BERT-style Masked Language Modeling (MLM) objective~\cite{BERT} to integratively learn from the 1D and 3Di sequences, enabling support for both sequence-level and structure-level tasks:
\begin{equation}
\mathcal{L}_{\text{1D-MLM}} = \mathbb{E}\left[\sum_{i\in\mathcal{M}}-\log p(s_i|\mathbf{s}_{/\mathcal{M}})\right],
\end{equation}
\begin{equation}
\mathcal{L}_{\text{3Di-MLM}} = \mathbb{E}\left[\sum_{i\in\mathcal{M}}-\log p(a_i|\mathbf{a}_{/\mathcal{M}})\right],
\end{equation}
where $\mathcal{M}$ is randomly masked positions and $\mathbf{s}_{/\mathcal{M}},\mathbf{a}_{/\mathcal{M}}$ denote the masked sequence where all masked positions have been replaced by the [MASK] token.

Specifically, for each 1D or 3Di sequence, 15\% of the corresponding tokens are randomly masked and then predicted based on the remaining contextualized representation. 
Consequently, the loss function $\mathcal{L}_\text{I}$ in pre-training stage I is formulated as follows, where $\mathcal{L}_{\text{1D-MLM}}$ stands for 1D sequences and $\mathcal{L}_{\text{3Di-MLM}}$ for 3Di sequences.
\begin{equation}
    \mathcal{L}_\text{I} = \mathcal{L}_{\text{1D-MLM}} + \mathcal{L}_{\text{3Di-MLM}}.
\end{equation}
% where $\mathcal{L}_{\text{1D-MLM}}$ stands for 1D sequences and $\mathcal{L}_{\text{3Di-MLM}}$ for 3Di sequences.

% simultaneously identify both 1D and 3Di sequence
% 引入蛋白数据辅助建模
Pre-training stage I endows the model with the capability to simultaneously identify both 1D and 3Di sequences. Furthermore, the efficient utilization of protein data in pre-training stage I effectively alleviates issues arising from insufficient antibody structure data. Due to the powerful generalization ability of large language models, the global structural constraints from proteins learned in stage I set the foundation for antibody specific learning of sub-domain local constraints in stage II.

\subsubsection{Stage II: Antibody Specific Multi-level Learning}
\label{sec:stage2}
% 1、阶段二的目的, 加结构的好处
% 2、SSM
% 3、CLR

% different granularities
% interdependency inherent in antibody sequences and structures
After pre-training stage I, \ourapproach has thoroughly comprehended 1D and 3Di sequences across the general protein domain. Subsequently in pre-training stage II, we can primarily focus on multi-level representation learning in the target antibody sub-domain. To better absorb comprehensive knowledge of antibody sequences and structures, exploring new pre-training mechanisms is worthwhile. Two multi-level learning objectives are introduced to inject different granularities of antibody specific sequential and structural information into an ALM: \textbf{Sequence-Structure Matching (SSM)} and \textbf{Cross-Level Reconstruction (CLR)}. The customized learning objectives facilitate the extraction of complex patterns and interdependency inherent in antibody sequences and structures.

Sequence-structure matching captures the coarse-grained alignment between antibody sequential and structural information. It is a binary classification task to predict whether a sequence-structure pair is matching or unmatching, as depicted in Fig.~\ref{Fig:framework}c. The model extracts representations of the corrupted antibody 1D and 3Di sequences, and then exploits a linear layer to make classification on their matching relationships.
SSM optimizes the following loss function:
\begin{equation}
    \mathcal{L}_{\text{SSM}} = \sum_{(\tilde{s},\tilde{a})\in\tilde{(S,A)}}H
    \left[y(\tilde{s},\tilde{a}), p(\mathcal{E}_{\theta}(\tilde{s},\tilde{a}))\right],
\end{equation}
where $\tilde{(S,A)}$ is the corrupted antibody dataset containing matched and unmatched sequence-structure pairs and $H$ denotes the cross entropy loss function. $\mathcal{E}_{\theta}$ and $y$ represent parameterized encoder and matching ground truth, respectively. 

Cross-level reconstruction focuses on improving fine-grained understanding in antibody sequence-structure pre-training, 
which differs in reconstruction conditions from MLM in Sec.~\ref{sec:stage1}. Concretely, the paired antibody 1D and 3Di sequences are separately encoded by the model. Then we randomly replace 15\% tokens with the [MASK] token in the 1D or 3Di sequence, whereas keeping the other level sequence unmasked. 
% interrelated mechanism
Eventually, as illustrated in Fig.~\ref{Fig:framework}d, hybrid information from both 1D and 3Di sequences serves as the cross-level reconstruction condition to make token-level predictions through a linear layer. CLR encourages the model to recover the corrupted 1D or 3Di sequences based on information from both levels, explicitly capturing the interrelated mechanism between antibody sequences and structures:
\begin{equation}
    \mathcal{L}_{\text{1D-CLR}} = \mathbb{E}\left[\sum_{i\in\mathcal{M}}-\log p(s_i|\mathbf{s}_{/\mathcal{M}};\mathbf{a})\right],
\end{equation}
\begin{equation}
    \mathcal{L}_{\text{3Di-CLR}} = \mathbb{E}\left[\sum_{i\in\mathcal{M}}-\log p(a_i|\mathbf{s};\mathbf{a}_{/\mathcal{M}})\right], 
\end{equation}
where $\mathcal{M}$ is randomly masked positions in 1D or 3Di sequences. $\mathcal{L}_{\text{1D-CLR}}$ and $\mathcal{L}_{\text{3Di-CLR}}$ subsequently measure the reconstruction costs of 1D and 3Di sequences.
Therefore, the ultimate loss function $\mathcal{L}_\text{II}$ in pre-training stage II is formulated as:
\begin{equation}
    \mathcal{L}_\text{II} =  \mathcal{L}_{\text{SSM}} + \mathcal{L}_{\text{1D-CLR}} + \mathcal{L}_{\text{3Di-CLR}}
\end{equation}
% where $\mathcal{L}_{\text{SSM}}$ and $\mathcal{L}_{\text{CLR}}$ are loss functions for sequence-structure matching and cross-level reconstruction respectively.

With the groundwork laid in pre-training stage I, the tailored pre-training objectives in stage II facilitate \ourapproach to effectively integrate antibody sequential and structural information, modeling comprehensive antibody representations. 
Overall, within this hierarchical pre-training paradigm, these two pre-training stages are complementary and indispensable to each other. Their synergy makes \ourapproach a powerful antibody foundation model, further fostering holistic antibody understanding and generation.
(See thorough pre-training details in Supplementary Materials Note S3.)

\section{Results}
\subsection{\ourapproach: pre-trained ALM integrating sequences and structures}
% large-scale & hybrid
In this paper, we pre-trained a large-scale antibody specific language model \ourapproach, which integrates multi-level information of 1D sequences and 3D structures for comprehensive antibody representation learning. 
% method
% general-purpose representations
To accomplish this, the model architecture is built upon ESM-2~\cite{esm-2}, comprising 650 million trainable parameters to facilitate the large-scale pre-training. Additionally, we collect a large-scale comprehensive training dataset, composed of sequences and structures from both protein and antibody domains. Furthermore, Foldseek~\cite{foldseek} is introduced to execute the efficient encoding of protein and antibody 3D structures, transforming 3D structures to 3Di sequences. Given sufficient 1D and 3Di sequences of proteins and antibodies, we propose a hierarchical pre-training paradigm, where two pre-training stages are constructed to foster holistic understandings of general proteins and specific antibodies correspondingly.
To further combine sequential and structural information, we develop two customized pre-training objectives, namely sequence-structure matching and cross-level reconstruction. Equipped with a comprehensive dataset, an efficient encoding technique, a hierarchical pre-training paradigm and tailored pre-training objectives, the resulting \ourapproach acquires general-purpose antibody representations containing biological information derived from both 1D sequences and 3D structures. 
% impact
% integration of sequence and structure encoding
% broadly benefit antibody understanding and generation applications
To our best knowledge, \ourapproach is the first large-scale pre-trained antibody specific language model that encodes sequences and structures in a hybrid manner for comprehensive representation learning.
% achieves multi-level integration of sequence and structure encoding 
In contrast to ALMs solely trained on massive sequences, our main contribution to the biological community lies in the additional infusion of structural information into \ourapproach. (See detailed advantages over other LLMs in Supplementary Materials Note S4.) The structure-enhanced representations demonstrate a strong potential to broadly benefit antibody understanding and generation applications.

\subsection{Observation of Multi-scale Organization in Antibody Representations}
% ESM-1b & BALM & SaProt & PALM
The variation observed in large antibody sequence datasets is influenced by processes at many scales, including functional properties that affect binding to specific antigens, species selection biases, and isotype biases that reflect different B cell maturation stages. Uncovering inherent patterns and properties is crucial for comprehending antibody properties. 
Unsupervised learning captures latent factors that, while unobserved, prove instrumental in elucidating the biological sequence variation~\cite{BALM}. 
We investigate the representation space of the pre-trained large language models at aforementioned multiple scales to look for signatures of biological organization. 

% t-SNE
% baselines
For a more intuitive comparison, t-SNE algorithm~\cite{t-SNE} is employed to visualize the antibody representations generated by the last layer of LLMs. To comprehensively evaluate the effectiveness of large-scale pre-training, it is necessary to compare representations before and after the pre-training phase~\cite{PALM}. We also include the well-established and powerful ESM-2~\cite{esm-2} (pre-trained on protein sequences) for comparison to further verify the superiority of the constructed \ourapproach.

\subsubsection{Pre-training Encodes Functional Specificity}
% data & impact
Antibodies have emerged as essential therapeutic agents in the treatment of various autoimmune, infectious and metabolic diseases, mainly owing to their ability to specifically bind to corresponding antigens~\cite{IgLM, reprogBERT}. Precise identification of antibody functional specificity will greatly enhance the progress of antibody development and optimization. From the OAS database~\cite{OAS}, we systematically identify and filter out 6,000 antibodies that exhibit specific binding affinities towards three distinct pathogens: HIV, Ebola virus and SARS-CoV-2.

\begin{figure}[!ht]
    \centering
    \includegraphics[width=0.98\linewidth]{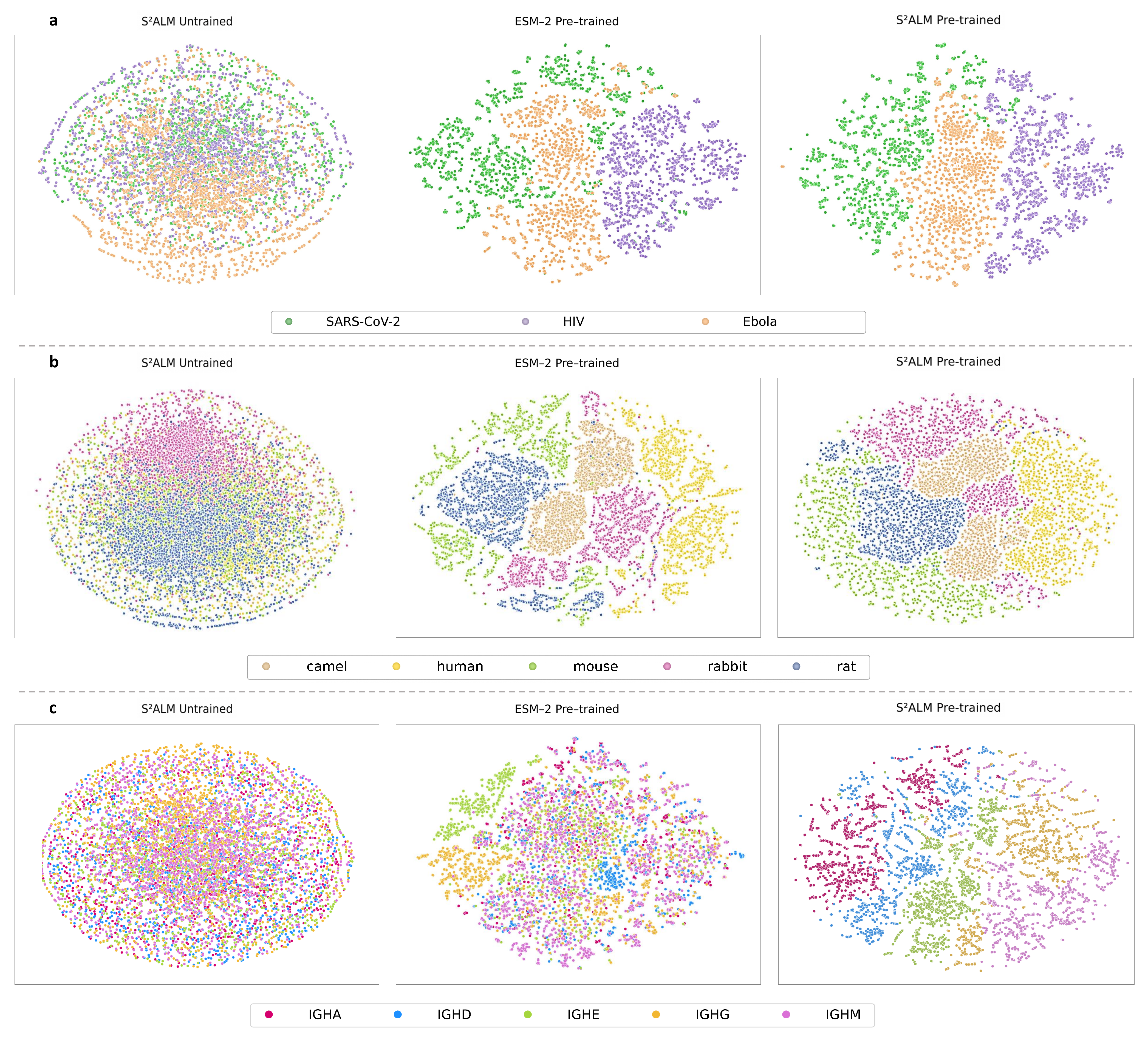}
    \caption{\textbf{The t-SNE visualization results.} Different colors indicate antibodies with different categories correspondingly. Untrained \ourapproach and pre-trained ESM-2 are included for comparison. The visualization analyses demonstrate that \ourapproach contains information about functional specificity, biological species and evolutionary isotypes in its comprehensive encoded representations.} 
    \label{Fig:t-SNE}
\end{figure}

% results
% confusion
We hereby present the t-SNE visualization analyses in Fig.~\ref{Fig:t-SNE}a. Surprisingly, ESM-2~\cite{esm-2} somewhat succeeds in distinguishing among these three types of antibodies, but there remains some confusion regarding antibodies targeting SARS-CoV-2 and the Ebola virus. The pre-trained \ourapproach produces more clearly aggregated antibody representations for all three pathogens, contrasting with scattered representations from the untrained model and ESM-2. This further verifies the necessity of our pre-training phase and indicates that the representations extracted by \ourapproach contain antibody functional specificity information.

\subsubsection{Pre-training Encodes Biological Species}
% data & impact
Due to significant differences in their genetic information and immune function mechanisms, antibodies from different species are expected to exhibit notable distribution disparities. We randomly sample and collect 15,000 antibody sequences from the OAS database~\cite{OAS}, ensuring an equal number of distinct species. The primary sources of biological species are taken into account for antibody sequencing, including camel, human, mouse, rabbit and rat.

% results
% confounding issue
The t-SNE visualization is shown in Fig.~\ref{Fig:t-SNE}b. We first observe that, when encoded by untrained \ourapproach, all antibodies from different species tend to mix together in the t-SNE two-dimensional space. Additionally, ESM-2~\cite{esm-2} successfully uncovers some clue about the species origins of various antibodies, whereas there are still some confounding issues regarding the boundaries of the clusters. Notably, antibodies of distinct species are nicely clustered together after being encoded by pre-trained \ourapproach, indicating the learned representations are encoded with rich biological species information.

\subsubsection{Pre-training Encodes Evolutionary Isotypes}
% data & impact
Antibodies of different isotypes activate distinct effector mechanisms, manifesting at different stages of the immune response, and differing in structures and locations. Isotype expression reflects the maturation stage of a B cell. Naive B cells express IgM and IgD isotypes with unmutated variable genes, while expression of other antibody isotypes (IgG, IgA, and IgE) occurs after antigen exposure~\cite{isotype}. Distinguishing antibody isotypes evaluates the injection of evolutionary information during model pre-training phase. We select 2,000 antibody sequences from the OAS database~\cite{OAS} per distinct isotype, resulting in a total of 10,000 sequences for t-SNE visualization analyses.

% results
As illustrated in Fig.~\ref{Fig:t-SNE}c, when exploiting representations extracted by untrained \ourapproach and the pre-trained ESM-2~\cite{esm-2} for t-SNE projection, the organizations of antibodies with various isotypes are predominantly diffuse. 
This may be attributed to the fact that isotype classification heavily relies on antibody specific evolutionary information, which untrained \ourapproach and the pre-trained ESM-2 are unable to capture. In contrast, we can observe a distinct clustering phenomenon utilizing antibody representations with different isotypes derived from pre-trained \ourapproach.
The visualization provides a sanity check for the ability of \ourapproach to extract the valuable information of antibody evolutionary isotypes.

\begin{figure}[!htbp]
    \centering
    \includegraphics[width=0.98\linewidth]{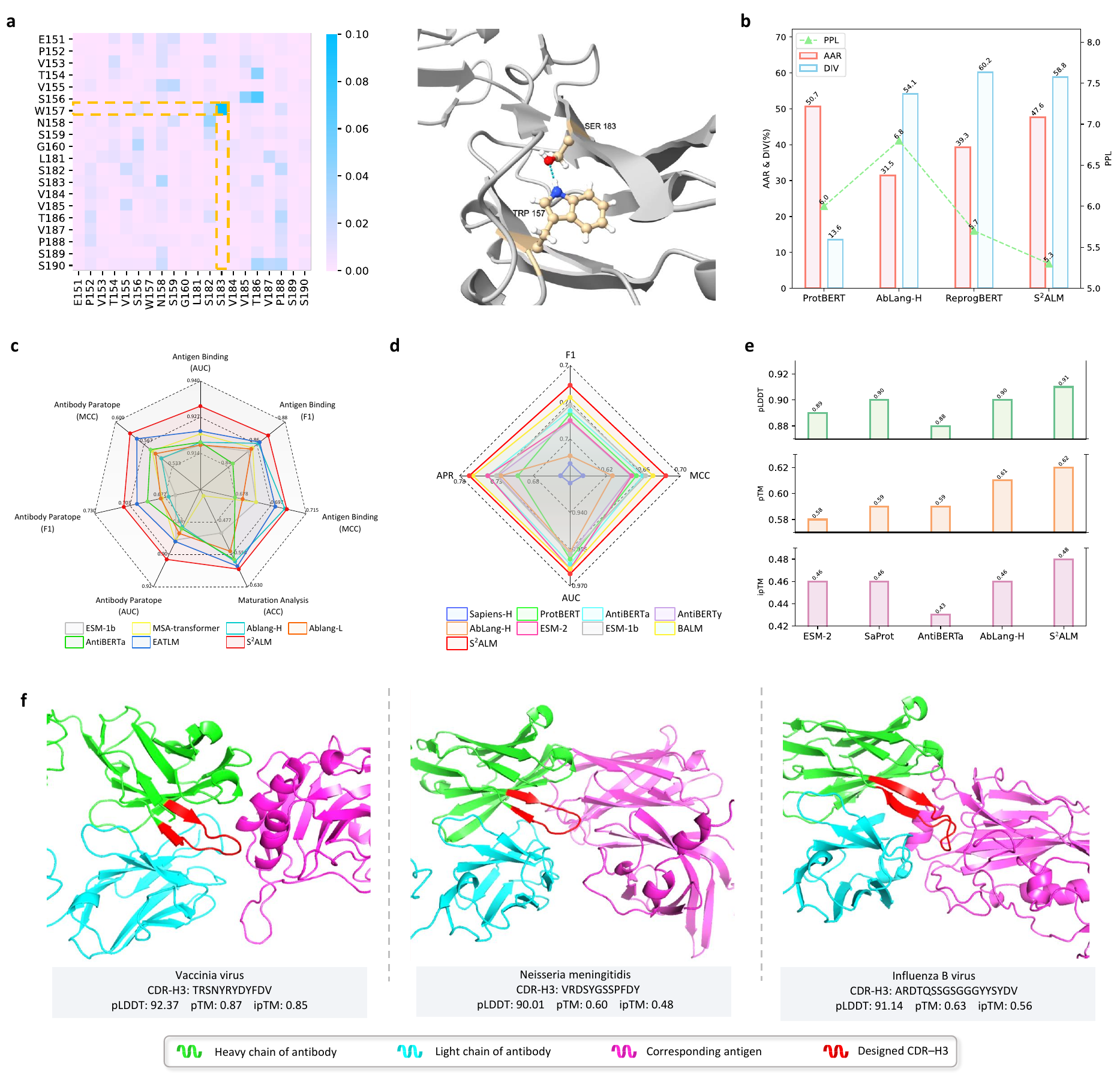}
    \caption{\textbf{\ourapproach exhibits superior performance on antibody understanding and generation tasks.} 
    \textbf{a}, Interpretability analysis of \ourapproach in capturing antibody structural interaction patterns.
    The heatmap reveals the self-attention values of the STE90-C11's heavy chain, derived from the last hidden layer of the 3th head in \ourapproach. The crystal structure of STE90-C11 (PDB: 7B3O) confirms the interaction mediated by the hydrogen bond between TRP157 and SER183.
    \textbf{b}, Evaluation results of the generated antibodies on antibody CDR design task. \ourapproach simultaneously balances the generative PPL, AAR and DIV.
    \textbf{c-d}, Experimental performance on antigen binding capability prediction, B cell maturation analysis and antibody paratope prediction tasks. 
    (Antibody paratope prediction datasets are from~\cite{EATLM} in \textbf{c} and~\cite{AntiBERTa} in \textbf{d}.). 
    \ourapproach consistently achieves state-of-the-art performance across all included evaluation metrics compared to all baseline models.
    \textbf{e}, Structural evaluation results of the generated antibodies. AlphaFold3 is employed to predict 3D structures. \ourapproach surpasses other baseline models in terms of pLDDT, pTM and ipTM.
    \textbf{f}, 3D structure visualization of the generated complexes. Targeting three specific pathogens (\textit{i.e.}, Vaccinia virus, Neisseria meningitidis and Influenza B virus), \ourapproach is employed to design the antibody CDR-H3 (highlighted in red). The stable and regular 3D structures of designed antigen-antibody complexes fully demonstrates the superiority of \ourapproach.}
    \label{Fig:statistics}
\end{figure}

\subsection{\ourapproach Captures Clues on Antibody Structure}
% impact
% ALMs identify and focus on crucial interaction sites
It is essential for antibody specific language models to identify and focus on crucial interaction sites during the representation learning process. And the multi-head self-attention mechanism, which specifically focuses on the different aspects of antibodies, has the potential to capture sophisticated interdependencies within the antibody structure.
% process
% interpretability analysis
% last hidden layer, 3th head
As a practical demonstration of \ourapproach's proficiency in modeling the antibody structural interaction patterns, we execute a structural interpretability analysis. And we find that residue pairs with high self-attention scores can accurately reveal long-range structural contacts. Specifically, the antibody STE90-C11 (PDB: 7B3O), a SARS-CoV-2 neutralizing antibody that binds to the ACE2-RBD interface, is selected as an example input. The self-attention scores from the last hidden layer in \ourapproach are extracted to build the heatmap. And we identify the potential hydrogen bond locations within the antibody structure. This allows assessing whether \ourapproach identifies antibody residue structural interactions.

% results
% substantiates the existence
As shown in Fig.~\ref{Fig:statistics}a, the heatmap displays a high self-attention score between residues TRP157 and SER183 (highlighted in orange), serving as an indicator of potential structural associations among these two residues. Correspondingly, the intricate crystal structure precisely substantiates the existence of a hydrogen bond between residues TRP157 and SER183.
% conclusion
% identify key interaction sites
While the interpretation of attention mechanisms remains an active area of research, this analysis provides an intuitive way to gain profound comprehension of the model's functional mechanism. \ourapproach showcases its ability to meaningfully highlight the key interaction sites within antibody structures, reaffirming the superiority of incorporating structural information during ALM pre-training.

% encompassing
\subsection{\ourapproach Accurately Predicts Antigen Binding Capacity}
% impact
Establishing the precise binding specificity between antibodies and their target antigens is central to accelerating advancements in therapeutic antibody optimization and deepening our comprehension of the intricacies inherent to the immune response. Consequently, rapid and accurate evaluation of the antigen binding capacity stands as a pressing and fundamental requirement in the realm of antibody research and development. 
% data
Antigen binding capacity prediction is a binary sequence classification task to determine whether the antibody can bind to the specific antigen, where the CDR-H3 region plays a pivotal role. With a keen focus on the interaction between the target antigen human epidermal growth factor receptor 2 (HER2) and the clinically approved wild-type antibody trastuzumab, we compile a dataset of antibody-expressing sequences. These sequences are designed by replacing the original trastuzumab sequence with a myriad of variant CDR-H3 fragments from the heavy chain, as detailed in~\cite{antigen_bind, EATLM}. The resulting dataset boasts an impressive scale of 21,612 unique antibody sequences and applies the training/validation/test split of 15,128/3,242/3,242 (\textit{i.e.}, 75\%/15\%/15\%). Our comparative analysis encompasses baselines of different types, including those pre-trained on protein sequences (\textit{i.e.}, ESM-1b~\cite{esm-1b}, MSA-1b~\cite{MSA-transformer}) and those pre-trained on antibody sequences (\textit{i.e.}, Ablang-H~\cite{AbLang}, Ablang-L~\cite{AbLang}, AntiBERTa~\cite{AntiBERTa}, EATLM~\cite{EATLM}). Three classification metrics are utilized for evaluation. AUC is the area under the receiver operating characteristic curve, which shows the performance at all classification thresholds. F1 is the average weighted score of precision and recall. Matthews correlation coefficient (MCC) is the coefficient between true and predicted values. 

% results
We report the evaluation results of the proposed \ourapproach and various baselines in Table~\ref{Tab:ATUE} (left). On the one hand, ESM-1b and MSA-1b yield relatively inferior performance due to the domain knowledge gap between proteins and antibodies. On the other hand, among baselines pre-trained on antibody sequences, Ablang-H outperforms Ablang-L and AntiBERTa, indicating the advantage of separate training for heavy and light chain sequences. However, the true standout in these evaluations is \ourapproach, which surpasses all baselines and sets a new state-of-the-art performance in this task. Such accomplishment underscores the capacity of \ourapproach to transcend traditional sequence-based models in antigen binding prediction.

\subsection{\ourapproach Precisely Distinguishes B Cell Maturation States}
% impact
B cells occupy a central position in the immune system's protective arsenal, due to their distinctive ability to generate antibodies. In the human body, these antibodies serve as a vital line of defense against invading pathogens~\cite{PALM}. Therefore, exploring and analyzing the process of B cell antibody maturation remains crucial, enhancing our understanding of the intricate mechanisms that unfold during immune system evolution (\textit{i.e.}, a critical biological process affecting the function and antigen binding specificity of antibodies)~\cite{Bcell_impact1, Bcell_impact2}.
% data
B cell maturation analysis is a 6-category classification task. Its core objective lies in accurately distinguishing the maturation states of B cell antibody sequences. Each antibody sequence belongs to one of \{\textit{immature, transitional, mature, plasmacytes, memory IgD+, memory IgD-}\} states. Accomplishing this necessitates a model capable of learning an evolution-aware representation sensitive to different B cell maturation states. 
We utilize a total of 88,094 antibody sequences from~\cite{Bcell_data, EATLM} with 6 maturation states, and apply the default data split. The evaluation metric is accuracy (ACC), calculating the ratio of correct predictions.

% results
Table~\ref{Tab:ATUE} (center) demonstrates the findings of model ability to distinguish between multiple B cell mature states. For the baselines pre-trained on protein sequences, both ESM-1b~\cite{esm-1b} and MSA-1b~\cite{MSA-transformer} present inferior performance. Even though armed with evolution-aware representations learned from multiple protein sequence alignments, MSA-1b performs poorly in this antibody maturation related task. Such phenomenon suggests that there is a gap between protein and antibody domains. For those pre-trained on antibody specific sequences (\textit{i.e.}, Ablang-H~\cite{AbLang}, Ablang-L~\cite{AbLang}, AntiBERTa~\cite{AntiBERTa}, EATLM~\cite{EATLM}), they deliver promising results in this task. Notably, \ourapproach, pre-trained on both antibody sequences and structures, achieves a substantial leap forward and surpasses all baselines by a large margin. The evaluation result highlights the efficacy of \ourapproach in improving the comprehension of B cell maturation and immune evolutionary mechanisms.

\begin{table}[htbp]
    \caption{Performance on antigen binding capacity prediction, B cell maturation analysis and antibody paratope prediction tasks. The \textbf{best} experimental results are highlighted in \textbf{bold}. \ourapproach achieves considerable improvements across all three antibody specific tasks. This reveals the potential of \ourapproach as the antibody specific foundation model constructed for comprehensive antibody representation learning.}
    \centering
    \setlength{\tabcolsep}{2.5em}
    \resizebox{1.0\textwidth}{!}{
    \Large
    \begin{tabular}{l|c c c|c|c c c}
            \toprule
            \makebox[0.17\textwidth][l]{\multirow{2}{*}{\textbf{Method}}} & \multicolumn{3}{c|}{\textbf{Antigen Binding Prediction}} & \multicolumn{1}{c|}{\textbf{Maturation Analysis}} & \multicolumn{3}{c}{\textbf{Antibody Paratope Prediction}} \\
            \cmidrule{2-8}
            & AUC & F1 & MCC & ACC & AUC & F1 & MCC \\
            \midrule
            \makebox[0.17\textwidth][l]{ESM-1b~\cite{esm-1b}} & 0.917 & 0.854 & 0.689 & 0.503 & 0.886 & 0.669 & 0.547 \\
            \makebox[0.17\textwidth][l]{MSA-1b~\cite{MSA-transformer}} & 0.921 & 0.857 & 0.689 & 0.416 & 0.887 & 0.679 & 0.557 \\
            \midrule
            \makebox[0.17\textwidth][l]{Ablang-H~\cite{AbLang}} & 0.918 & 0.861 & 0.704 & 0.570 & 0.878 & 0.674 & 0.546 \\
            \makebox[0.17\textwidth][l]{Ablang-L~\cite{AbLang}} & 0.917 & 0.856 & 0.682 & 0.546 & 0.882 & 0.680 & 0.553 \\
            \makebox[0.17\textwidth][l]{AntiBERTa~\cite{AntiBERTa}} & 0.918 & 0.843 & 0.678 & 0.565 & 0.879 & 0.690 & 0.559 \\
            \makebox[0.17\textwidth][l]{EATLM~\cite{EATLM}} & 0.922 & 0.862 & 0.699 & 0.581 & 0.887 & 0.698 & 0.575 \\
            \midrule
            \makebox[0.17\textwidth][l]{\ourapproach} & \cellcolor{gray!50}\textbf{0.931} & \cellcolor{gray!50}\textbf{0.868} & \cellcolor{gray!50}\textbf{0.705} & \cellcolor{gray!50}\textbf{0.588} & \cellcolor{gray!50}\textbf{0.898} & \cellcolor{gray!50}\textbf{0.708} & \cellcolor{gray!50}\textbf{0.583} \\
            \bottomrule
    \end{tabular}}
    \label{Tab:ATUE}
\end{table}

\subsection{\ourapproach Thoroughly Identifies Antibody Paratopes}
% impact
In immunology research, identifying the amino acids in immunoglobulins that specifically bind to antigens (\textit{i.e.}, the antibody paratope) is one of the most crucial and challenging tasks~\cite{BCR}. The antibody paratope, usually composed of multiple amino acids in the complementarity-determining regions, forms the antibody binding site which is essential for antigen interactions in the immune response~\cite{Paratome}. Antibody paratope prediction could significantly help to enhance our understanding of the binding mechanisms of therapeutic antibodies.
% data
It is a binary sequence labeling task that identifies specific binding positions within antibody sequences, assigning a 0/1 label to each amino acid residue. 
For paratope prediction, we utilize two antibody datasets sourced from~\cite{EATLM, AntiBERTa}, which comprise 277 and 900 antibody sequences annotated with token-wise paratope regions. 
Numerous baseline models are utilized, including Sapiens-H~\cite{Sapiens}, ProtBERT~\cite{ProtTrans}, AntiBERTa~\cite{AntiBERTa}, AntiBERTy~\cite{AntiBERTy}, ESM-1b~\cite{esm-1b}, ESM-2~\cite{esm-2}, BALM~\cite{BALM}, MSA-1b~\cite{MSA-transformer} Ablang-H~\cite{AbLang}, Ablang-L~\cite{AbLang}, and EATLM~\cite{EATLM}.
Experimental metrics used to evaluate the prediction capability include area under the receiver operating characteristic curve (AUC), F1 score and Matthews correlation coefficient (MCC).

% results
Table~\ref{Tab:ATUE} (right) reports results on antibody paratope prediction benchmark~\cite{EATLM}, and we observe that \ourapproach achieves considerable improvements while other baselines show no significant difference in performance. 
Fig.~\ref{Fig:statistics}d depicts results on antibody paratope prediction benchmark~\cite{AntiBERTa}, where \ourapproach stands out by outperforming all other baselines across all criteria. 
The holistic experimental results showcase the superiority of our model in efficiently capturing features associated with antibody binding sites. This further highlights \ourapproach's high specificity, accuracy, and in-depth understanding of antibody complex binding mechanisms.

% The proposed hierarchical pre-training paradigm allows \ourapproach to excel in prediction of the antibody paratope.

\begin{table}[htbp]
    \caption{The performance comparison on antigen-antibody binding affinity prediction tasks. We highlight the \textbf{best} experimental results in \textbf{bold}. \ourapproach achieves state-of-the-art performance across all evaluation metrics on the three antibody datasets (\textit{i.e.}, 14H, 14L and BioMap).}
    \centering
    \setlength{\tabcolsep}{0.8em}
    \resizebox{1.0\textwidth}{!}{
    \Large
    \begin{tabular}{l|c c|c c|c c}
            \toprule
            \multirow{2}{*}{\textbf{Method}} & \multicolumn{2}{c|}{\textbf{14H}} & \multicolumn{2}{c|}{\textbf{14L}} & \multicolumn{2}{c}{\textbf{BioMap}} \\
            \cmidrule{2-7}
             & Pearson & Spearman & Pearson & Spearman  & Pearson & Spearman  \\
            \midrule
            Vanilla BERT~\cite{BERT} & 0.594 (0.021) & 0.480 (0.025) & 0.607 (0.024) & 0.611 (0.027) & 0.492 (0.036) & 0.498 (0.037) \\
            \midrule
            Ens-Grad~\cite{Ens-Grad} & 0.601 (0.016) & 0.476 (0.023) & 0.637 (0.019) & 0.645 (0.023) & 0.645 (0.031) & 0.664 (0.033) \\
            \midrule
            AbMAP~\cite{AbMAP} & 0.606 (0.015) & 0.510 (0.015) & 0.674 (0.012) & 0.685 (0.016) & 0.637 (0.027) & 0.673 (0.029) \\
            \midrule
            AntiBERTa2~\cite{AntiBERTa2} & 0.623 (0.011) & 0.545 (0.008) & 0.673 (0.013) & 0.684 (0.012) & 0.633 (0.022) & 0.670 (0.031) \\
            \midrule
            ESM-F~\cite{esm-2} & 0.634 (0.007) & 0.516 (0.010) & 0.674 (0.011) & 0.681 (0.014) & 0.628 (0.028) & 0.644 (0.024) \\
            \midrule
            A2binder~\cite{PALM} & 0.642 (0.012) & 0.553 (0.011) & 0.683 (0.010) & 0.688 (0.015) & \textbf{0.701 (0.024)} & 0.746 (0.025) \\
            \midrule
            \ourapproach & \cellcolor{gray!50}\textbf{0.650 (0.013)} & \cellcolor{gray!50}\textbf{0.558 (0.014)} & \cellcolor{gray!50}\textbf{0.687 (0.015)} & \cellcolor{gray!50}\textbf{0.693 (0.017)} & \cellcolor{gray!50}\textbf{0.701 (0.022)} & \cellcolor{gray!50}\textbf{0.749 (0.028)} \\
            \bottomrule
    \end{tabular}}
    \label{Tab:binding_affinity}
\end{table}

\subsection{\ourapproach Accurately Predicts Antigen-Antibody Binding Affinity}
% impact
% therapeutic antibody development
Therapeutic antibody development has become an increasingly popular strategy for drug discovery, but still involves intensive laboratory experiments~\cite{Pfam_BERT}. Recently, deep learning methods have emerged as a promising approach to accelerate such process by identifying high-affinity antibody candidates~\cite{Pfam_BERT, IgT5}. The property of binding affinity, which quantifies the binding strength between antigen-antibody pairs, plays a critical role in antibody development and optimization.
% data
To assess the model's performance in predicting the binding affinity, we first utilize two datasets, 14H and 14L~\cite{14HL}, which contain abundant antibodies labeled with binding affinity values. Within these two datasets, the affinity values quantify the binding strength of antibodies to a stable peptide in the HR2 region of SARS-CoV-2. To further expand the diversity of binding antigens and thereby enhance both the comprehensiveness and reliability of this experiment, we also include the BioMap~\cite{BioMap} dataset for the evaluation of binding affinity prediction. The BioMap dataset contains 1,706 antigen-antibody pairing data with labeled binding affinity values, in which most antibodies stem from human and mouse sources. As for the experiment execution, the 14H, 14L and BioMap datasets are each divided into training set, validation set and test set according to the split of 80\%/10\%/10\% following~\cite{PALM}. Based on the test sets, we record experimental results for fair comparison with extensive baselines (\textit{i.e.}, Vanilla BERT~\cite{BERT}, Ens-Grad~\cite{Ens-Grad}, AbMAP~\cite{AbMAP}, AntiBERTa2~\cite{AntiBERTa2}, ESM-F~\cite{esm-2}, and A2binder~\cite{PALM}).  In this binding affinity prediction task, we use the Pearson correlation and the Spearman correlation as evaluation metrics.
% model architecture
Building upon pre-trained \ourapproach, we extend and enhance the model architecture to enable the accurate learning of antigen-antibody binding rules. Concretely, \ourapproach is utilized to extract features of antibodies, while ESM-2~\cite{esm-2} is introduced for the feature encoding of antigens. Following each language model is a Multi-Fusion Convolutional Neural Network (MF-CNN), which is designed to combine the antigen and antibody features extracted by pre-trained models. Finally, the MLP (a multilayer perceptron) model fuses both features of antigen and antibody to predict the binding affinity.

% results
Table~\ref{Tab:binding_affinity} illustrates the performance comparison of models in predicting the antigen-antibody binding affinity. \ourapproach consistently achieves state-of-the-art performance in Pearson correlation and Spearman correlation metrics. And we observe that \ourapproach significantly surpasses all baselines, except for the BioMap dataset, where it achieves the same Spearman correlation as A2binder. The experimental results further verify the heightened precision of \ourapproach in predicting the antigen-antibody binding affinity. By integrating both sequence and structure data for large-scale pre-training, \ourapproach effectively captures the specific functional information of antibodies. We believe that \ourapproach has the great potential to serve as the antibody specific foundation model to promote the therapeutic antibody development in real-world scenarios.

\subsection{\ourapproach Enables Computational Antibody CDR Design}
\label{CDR_design}
% impact
Given the extraordinary performance on antibody understanding tasks related to the evolutionary process and functional mechanism, we additionally incorporate a novel generation task to further evaluate model's capability of antibody generation and optimization. The Complementarity-Determining Region (CDR), also known as the hypervariable region, primarily determines the antibody’s affinity and specificity for its target. Given the crucial role of CDRs, computational antibody design primarily focuses on generating these regions.
% data
% 1D sequence + 3Di sequence
Specifically, antibody CDR design is a sequence infilling task to generate masked CDRs (more exactly, CDR-H3) based on the contextualized representation. 
% Note that CDRs are complementarity determining regions in the variable domains of an antibody and play their significant roles by binding to specific antigens. 
For antibody CDR design, the dataset curated by~\cite{refine-GNN} follows the training/validation/test split of 2,282/291/291 samples derived from the Coronavirus Antibody Database (CoV-AbDab)~\cite{CoV_AbDab}. CoV-AbDab is a public database documenting molecular information (sequences and structures) of all published and patented antibodies and nanobodies that can bind to coronaviruses, including SARS-CoV-2, SARS-CoV-1 and MERS-CoV~\cite{CoV_AbDab}. 
% experimental details
During this experiment, we mask the CDRs in both 1D and 3Di antibody sequences for \ourapproach to reconstruct. 
To benchmark the quality of antibody sequences generated by \ourapproach, we employ ProtBERT~\cite{ProtTrans}, AbLang-H~\cite{AbLang} and ReprogBERT~\cite{reprogBERT} for comparison.
And three metrics are utilized for holistic evaluation. Perplexity (PPL) measures how well the language model predicts the generative tokens. Here we exploit off-the-shelf ProGen~\cite{progen} for calculation following~\cite{reprogBERT}. Lower values signify better performance, indicating stronger ``naturalness'' of the generated CDRs. Amino acid recovery (AAR) is computed for the specific sequence region of interest, measuring the percent of the exact matches between ground truth and sampled sequences. The higher the AAR, the more accurate the recovery. Diversity (DIV) computes the complement of the average recovery of all pairwise comparisons using sampled CDRs. A higher value corresponds to an increased dissimilarity among the samples themselves. 

% results
% antibody design
The evaluation results are shown in Fig.~\ref{Fig:statistics}b. In spite of the high AAR, ProtBERT struggles in low generation diversity. AbLang-H and ReprogBERT are particularly talented in generation diversity, but demonstrate relatively high perplexity. In contrast, \ourapproach achieves substantially high AAR and the lowest perplexity while maintaining the comparable high diversity to ReprogBERT, accomplishing the balance of all three generative metrics. Leveraging the CDR infilling paradigm, \ourapproach successfully generates CDR loop libraries with improved in silico developability profiles. This strongly indicates that \ourapproach could serve as an effective and insightful tool for computational antibody design, benefiting the biomedical research community.

\subsection{3D Structural Visualization}
\label{visualization}
To further manifest the advantages of incorporating structural information during ALM pre-training, we conduct 3D visualization of antigen-antibody complexes generated by \ourapproach. Based on a holistic antigen-antibody complex dataset of 60 antigen-antibody pairs constructed by~\cite{RosettaAbDesign}, \ourapproach designs antibody CDR-H3 via sequence infilling. For each antigen-antibody complex, 100 CDR-H3 samples are generated and combined with standard framework regions to obtain 100 full-length antibody heavy chains. 
% ALphaFold3
% follow the standard protocol
We select the top-3 antibody heavy chains with the maximum naturalness using ProGen2~\cite{progen2}, following the standard protocol~\cite{reprogBERT, PALM}. Subsequently, AlphaFold3~\cite{alphafold3}, highly capable of predicting the antigen-antibody joint structure, is applied to generate the 3D structures of complexes comprising antigen and generated antibodies (including antibody heavy chains and corresponding light chains). 
% metric and baseline
% predicted confidence
Evaluation metrics of the folded 3D structure are the predicted template-modeling score (pTM), the interface pTM (ipTM), and the predicted local distance difference test (pLDDT). Moreover, to facilitate a thorough comparison, we involve a variety of language models: ESM-2~\cite{esm-2}, pre-trained on general protein sequences; SaProt~\cite{SaProt}, which harnesses both protein sequences and structures for pre-training; and two dedicated ALMs, AntiBERTa~\cite{AntiBERTa} and AbLang-H~\cite{AbLang}.

% results
The structural evaluation results through AlphaFold3 are illustrated in Fig.~\ref{Fig:statistics}e. SaProt and AbLang-H stand out among all baselines, verifying the advantages of injecting structural information and promoting antibody specific learning. \ourapproach consistently achieves the optimal pTM, ipTM, and pLDDT scores compared to other methods, demonstrating the generated antibodies are more likely to target the corresponding antigens and form stable binding complexes.
% vis
Furthermore, we visualize the generated 3D complex structures targeting three clinically significant pathogens (\textit{i.e.}, Vaccinia virus, Neisseria meningitidis and Influenza B virus). Fig.~\ref{Fig:statistics}f depicts some visualization examples. The stable and regular 3D structures of designed antigen-antibody complexes fully exhibit the superior generative capability of \ourapproach, which pioneeringly incorporates 3D structural information into ALM pre-training.

\section{Discussion}
% background
Antibodies are vital proteins produced by the immune system to safeguard against and combat with a variety of diseases~\cite{antibody_1}. There are abundant clinical applications of antibody analysis in therapeutic drug discovery and finding novel diagnostics~\cite{AntiBERTa, IgLM, PALM, ProtET}. Most analyses have focused on comparing high-level differences between cohorts to facilitate antibody understanding, such as number of somatic hypermutations, isotype subclass usage, and V(D)J gene segment usage. With the advancements of machine learning techniques, it has been verified in various studies that transformer-based LLMs significantly promote the understanding and decoding of biological language~\cite{Evo, IgLM, PALM, Bridge-IF, HDMLF}. 
The biological function of an antibody is directly determined by its 3D structure, so the modalities of sequence and structure should be tightly coupled to generate a comprehensive representation of the antibody.
% The comprehensive representation primarily originates from different levels of antibodies: one-dimensional amino acid sequence and three-dimensional natural folded structure. 
% Given the indispensable role of 3D structures directly determining the binding functionality, 
% Although 3D structures play an indispensable role in directly determining the binding functionality, 
However, existing works fall short in effectively injecting structural information during antibody pre-training, which significantly hinders the further progress.

% findings
In this study, our endeavor herein offers an antibody specific foundation model that posits the potential to facilitate comprehensive antibody understanding and generation. Extensive machine learning paradigms and techniques are utilized in developing \ourapproach. 
To our best knowledge, \ourapproach stands out as the first ALM to integrate hybrid information of 1D sequences and 3D structures during large-scale pre-training. 
\ourapproach extracts comprehensive and meaningful antibody representations, uncovering inherent patterns and properties of functional specificity, biological species and evolutionary isotypes. Additionally, as a supplement to interpretability analysis, \ourapproach precisely captures crucial interaction sites within 3D structures.
Evaluated on diverse computational scenarios, \ourapproach exhibits the state-of-the-art performance in antibody representation learning: accurately predicting antigen binding capacity, precisely distinguishing B cell maturation states, thoroughly identifying antibody paratopes, and accurately predicting antigen-antibody binding affinity. To further explore the potential applications, we employ \ourapproach for computational antibody design leveraging the CDR infilling paradigm. The generated stable and regular 3D structures further indicate the superior generative capability of \ourapproach.
We hold the expectation that our work will usher in a new era of antibody multi-level pre-training via explicitly modeling the profound interrelation between sequences and structures, fostering the development of therapeutic antibody drugs in real-world scenarios.

% limitations and broader impacts
% params & data & Foldseek
While our work demonstrates promising and outstanding results in extensive antibody specific downstream tasks, there are still some areas for improvement and future exploration: (1) Due to computational constraints, the model size of \ourapproach may not have reached its maximum capacity. (2) Although we circumnavigate such constraints by introducing extra protein structures in our hierarchical pre-training paradigm, the limitation of antibody structure data cannot be overlooked. In the future, we eagerly anticipate the emergence of large-scale antibody structure databases with sufficient experimentally-determined 3D structures. Such efforts will fill the void of large-scale antibody structure corpus, thereby fueling multi-level antibody pre-training. (3) The performance of structural pre-training heavily depends on Foldseek~\cite{foldseek}, which aims to balance search efficiency and encoding accuracy. Thus there is still room for improving the representation capability of all methods building upon Foldseek, including ours.

% impact
To conclude, this paper documents our efforts to build a comprehensive antibody specific large language model to represent the intricacies of the biological world. The capabilities demonstrated by \ourapproach exhibit considerable promise and underscore several areas in antibody understanding and generation that necessitate substantial advancements. Such a multi-level pre-trained foundational model, integrating antibody sequential and structural information, will prove immensely valuable in accelerating immunotherapy development and enhancing our comprehension of biological phenomena.

\section*{Acknowledgments}
We sincerely thank the Department of Medicine, Zhejiang University and the Department of Computer Science, Zhejiang University for their support.

\subsection*{Funding}
This research was partially supported by National Natural Science Foundation of China (62106218, 62176231, 82202984 and 12326612); Zhejiang Key R\&D Program of China (2023C03053,  2024SSYS0026 and 2024C03048); the Opening Foundation of the State Key Laboratory of Transvascular Implantation Devices (SKLTID2024003).

\subsection*{Author Contributions} 
M. Yin, H. Zhou and J. Chen conceived the idea. M. Yin, H. Zhou and Y. Zhu developed the methodology and designed the experiments. M. Yin, H. Zhou and Z. Kong collected the data. M. Yin, Y. Zhan conducted the experiments. Y. Zhan, J. L. Wu and H. Xu conducted the visualization. M. Yin, H. Zhou and J. L. Wu wrote the manuscript. C. Hsieh, J. Chen and T. Hou revised the manuscript. T. Hou and J. Wu supervised the study. All authors revised and approved the final version of the manuscript.

\subsection*{Conflicts of Interest}
The authors declare that they have no competing interests.

\subsection*{Data Availability}
The curated biotext-protein paired dataset and the source code to perform CLIP-informed protein editing will both be freely open-source when the paper is accepted. 

\appendix
\section*{Supplementary Materials}
\renewcommand{\thefigure}{S\arabic{figure}}
\setcounter{figure}{0}
\renewcommand{\thetable}{S\arabic{table}}
\setcounter{table}{0}
Notes S1 to S3 \\
Fig. S1 to S2 \\
Table S1 to S2

\subsection*{Note S1. Research Background}
% pre-training data organization
% efficient encoding approach
Utilizing massive amounts of text sequences, the transformer-based large-scale pre-training has become a widely adopted paradigm that demonstrates its remarkable capabilities in the field of Natural Language Processing (NLP)~\cite{BERT, GPT-4}. Similar in the realm of life sciences, protein language models (PLMs) and antibody language models (ALMs), pre-trained on extensive biomolecular sequences, have showcased outstanding performance across a diverse array of tasks related to biological structures and functions~\cite{esm-1b, esm-2, AntiBERTa, IgLM, EATLM}. Leveraging 1D sequential information, these PLMs and ALMs execute large-scale self-supervised pre-training to learn the language of life. However, since the three-dimensional spatial structures are directly relevant to biological functions, an intriguing and promising direction for their development is to incorporate extra 3D structure information into large-scale pre-training. Additionally, there have been some notable attempts in the protein domain but few in the antibody domain. It is primarily due to two critical challenges: (1) how to organize integrative pre-training data; (2) lack of efficient structural encoding technique. In this paper, we pay high attention to these two challenges and overcome them individually to pave the way for the training of \ourapproach. A holistic pre-training dataset is constructed, encompassing various levels and spanning multiple domains. And Foldseek in introduced to accomplish the efficient 3D structure encoding.
% since the traditional transformer architecture cannot directly encode 3D spatial structures, exploring new techniques to encompasses both sequential and structural information is worthwhile.

\subsection*{Note S2. Motivations of \ourapproach}
% 之前的人没考虑到抗体结构，结构很重要，所以我们是第一个加入抗体结构做预训练的。但是抗体结构数据欠缺，所以加入蛋白的序列和结构数据。
Existing ALMs learn antibody representations primarily based on their 1D residue sequences. However, the 3D structures encompass intricate spatial geometric knowledge and are directly relevant to antibody functions, which demonstrates the full potential to enhance antibody representation learning. In this paper, we organize pre-training data including antibody sequences and structures and propose to integrate sequential and structural information during pre-training. Since the scale of available antibody structure data is relatively small compared to the protein data, the additional protein data comprising sequences and structures is introduced as a compensation, assisting the ALM to comprehensively incorporate structural information. By taking advantage of information from multiple levels and additional domains, we execute hierarchical pre-training to obtain an ALM effective on various antibody specific downstream tasks.

\subsection*{Note S3. Pre-training Details}
% architecture
Building on the architecture of ESM-2~\cite{esm-2}, \ourapproach has 650 million trainable parameters. We incorporate 33 transformer encoder blocks, each containing 20 self-attention heads. And the hidden dimension is set to 1280. 
We exploit layer normalization to stabilize and speed up the training process by reducing internal covariate shift.
% positional encoding
The Rotary Position Embedding (RoPE) is employed to supply token positional information. Additionally, we truncate all training sequences to a maximum length of 1024. Sequences of length less than 1024 are padded, and padded tokens are excluded from the loss computation. 
% type encoding
By utilizing token-type encoding, the model with mixed training distinguishes 1D and 3Di sequences, assigning 0 to 1D sequences and 1 to 3Di sequences.

% objectives
Following BERT~\cite{BERT} and ESM-2~\cite{esm-2}, during training of Masked Language Modeling (MLM), 15\% of tokens in each batch are randomly selected and masked. For these selected tokens, 80\% are substituted with the [MASK] special token, while 10\% of them are replaced by random amino acids and the remaining 10\% are left unchanged. 
% in-batch sampling strategy
For the training of Sequence-Structure Matching (SSM), we use an in-batch sampling strategy similar to CLIP~\cite{clip} to corrupt the pairs of 1D and 3Di sequences. 
Given a mini-batch with $N$ paired data, we construct an ensemble of $N^2$ 1D and 3Di sequence pairs (including $N$ correct pairings and $N^2 - N$ incorrect pairings) for model to predict whether they match or not.
% the 1D or 3Di fragment
Regarding Cross-Level Reconstruction (CLR), we aim to reconstruct single-level information (\textit{i.e.}, solely 1D sequence or 3Di sequence) based on both levels of information.
% MLM
Note that the MLM task on antibody 1D sequences is additionally interspersed in pre-training stage II. 
Such operation avoids catastrophic forgetting and ensures the preservation of sequential information when incorporating cross-level insights from antibody structures.

% pre-training setup
Our model is implemented using the PyTorch framework~\cite{pytorch} and pre-trained on 32 NVIDIA Tesla V100 GPUs. For large-scale distributed training, we exploit the DeepSpeed ZeRO Stage 2 strategy~\cite{zero}.
The learning rate is $4 \times 10^{-4}$ and $1 \times 10^{-3}$ for pre-training stage I and II respectively. We also leverage mixed precision training for efficient pre-training.
% AdamW (huggingface Trainer default)
For training optimization, an AdamW optimizer is used with weight decay of 0.01. And we set a linear learning rate scheduler with 2,000 warmup steps.
% open-access
Eventually, we will make our codes, model weights, and the associated datasets openly available upon acceptance. These materials are expected to be valuable for both the computational and biological communities.

\begin{table*}[!ht]
    % the first ALM to simultaneously incorporate protein sequences, antibody sequences, protein structures, antibody structures for large-scale pre-training
    % protein: 序列+结构; antibody:蛋白+抗体序列。 
    \caption{Different LLMs, pre-trained for protein and antibody representation learning, vary in their utilization of distinct pre-training data types, objectives and downstream tasks. To the best of our knowledge, \ourapproach is the first ALM to simultaneously incorporate protein sequences, antibody sequences, protein structures, antibody structures for large-scale pre-training, accomplishing comprehensive antibody representation learning. Additionally, two novel pre-training objectives are proposed to integrate sequential and structural information. Furthermore, the superiority of \ourapproach is holistically assessed on a variety of antibody-related experiments. CLM: causal language modeling; MLM: masked language modeling. CRD: coordinate prediction; ILM: infilling language modeling; AGP: ancestor germline prediction; MPP: mutation position prediction. SSM: sequence-structure matching; CLR: cross-level reconstruction.}
    \centering
    \resizebox{1.0\textwidth}{!}{
    \begin{tabular}{l|c|c c c c|c|c}
            \toprule
            \multirow{3}{*}{} &\multirow{3}{*}{}& \multicolumn{4}{c|}{\textbf{Pre-training data}} & \multirow{3}{*}{} & \multirow{3}{*}{}\\
            \cmidrule{3-6}
            \multicolumn{1}{c|}{\textbf{Model}} & \textbf{LLM category} & \multicolumn{2}{c|}{Sequence} & \multicolumn{2}{c|}{Structure} & \textbf{Objective} & \textbf{Antibody specific downstream task}\\
            \cmidrule{3-6}
             & & Protein & \multicolumn{1}{c|}{Antibody} & Protein & Antibody & & \\
            \midrule
            ProtBERT~\cite{ProtTrans} & PLM & \textcolor{ForestGreen}{\usym{2713}} & \textcolor{red}{\usym{2717}} & \textcolor{red}{\usym{2717}} & \textcolor{red}{\usym{2717}} & MLM & - \\
            ESM-1b~\cite{esm-1b} & PLM & \textcolor{ForestGreen}{\usym{2713}} & \textcolor{red}{\usym{2717}} & \textcolor{red}{\usym{2717}} & \textcolor{red}{\usym{2717}} & MLM & - \\
            ESM-2~\cite{esm-2} & PLM & \textcolor{ForestGreen}{\usym{2713}} & \textcolor{red}{\usym{2717}} & \textcolor{red}{\usym{2717}} & \textcolor{red}{\usym{2717}} & MLM & - \\
            ProGen~\cite{progen} & PLM & \textcolor{ForestGreen}{\usym{2713}} & \textcolor{red}{\usym{2717}} & \textcolor{red}{\usym{2717}} & \textcolor{red}{\usym{2717}} & CLM & - \\
            PromptProtein~\cite{PromptProtein} & PLM & \textcolor{ForestGreen}{\usym{2713}} & \textcolor{red}{\usym{2717}} & \textcolor{ForestGreen}{\usym{2713}} & \textcolor{red}{\usym{2717}} & MLM+CRD & - \\
            SaProt~\cite{SaProt} & PLM & \textcolor{ForestGreen}{\usym{2713}} & \textcolor{red}{\usym{2717}} & \textcolor{ForestGreen}{\usym{2713}} & \textcolor{red}{\usym{2717}} & MLM & - \\
            \midrule
            AntiBERTa~\cite{AntiBERTa} & ALM & \textcolor{red}{\usym{2717}} & \textcolor{ForestGreen}{\usym{2713}} & \textcolor{red}{\usym{2717}} & \textcolor{red}{\usym{2717}} & MLM & functionality \\
            AntiBERTy~\cite{AntiBERTy} & ALM & \textcolor{red}{\usym{2717}} & \textcolor{ForestGreen}{\usym{2713}} & \textcolor{red}{\usym{2717}} & \textcolor{red}{\usym{2717}} & MLM & functionality \& evolution \\
            AbLang~\cite{AbLang} & ALM & \textcolor{red}{\usym{2717}} & \textcolor{ForestGreen}{\usym{2713}} & \textcolor{red}{\usym{2717}} & \textcolor{red}{\usym{2717}} & MLM & generation \\
            IgLM~\cite{IgLM} & ALM & \textcolor{red}{\usym{2717}} & \textcolor{ForestGreen}{\usym{2713}} & \textcolor{red}{\usym{2717}} & \textcolor{red}{\usym{2717}} & ILM & functionality \& generation \\
            PALM\cite{PALM} & ALM & \textcolor{red}{\usym{2717}} & \textcolor{ForestGreen}{\usym{2713}} & \textcolor{red}{\usym{2717}} & \textcolor{red}{\usym{2717}} & MLM+CLM & generation \\
            EATLM~\cite{EATLM} & ALM & \textcolor{red}{\usym{2717}} & \textcolor{ForestGreen}{\usym{2713}} & \textcolor{red}{\usym{2717}} & \textcolor{red}{\usym{2717}} & MLM+AGP+MPP & functionality \& evolution \\
            IgBERT, IgT5~\cite{IgT5} & ALM & \textcolor{ForestGreen}{\usym{2713}} & \textcolor{ForestGreen}{\usym{2713}} & \textcolor{red}{\usym{2717}} & \textcolor{red}{\usym{2717}} & MLM & functionality \\
            \midrule
            \ourapproach stage I & ALM & \textcolor{ForestGreen}{\usym{2713}} & \textcolor{red}{\usym{2717}}  & \textcolor{ForestGreen}{\usym{2713}} & \textcolor{red}{\usym{2717}} & MLM & \\
            \ourapproach stage II & ALM & \textcolor{red}{\usym{2717}} & \textcolor{ForestGreen}{\usym{2713}} & \textcolor{red}{\usym{2717}} & \textcolor{ForestGreen}{\usym{2713}} & MLM+SSM+CLR & functionality \& evolution \& generation\\
            \ourapproach & ALM & \textcolor{ForestGreen}{\usym{2713}} & \textcolor{ForestGreen}{\usym{2713}} & \textcolor{ForestGreen}{\usym{2713}} & \textcolor{ForestGreen}{\usym{2713}} & MLM+SSM+CLR &  \\
            \bottomrule
    \end{tabular}}
    \label{Tab:comparison}
\end{table*}

\subsection*{Note S4. Advantages over other LLMs}
Table~\ref{Tab:comparison} presents distinct pre-training data types, objectives, downstream tasks. Earlier PLMs~\cite{ProtTrans, esm-1b, esm-2, progen} exclusively focus on protein sequences, which is similar to current ALMs~\cite{AntiBERTa, AntiBERTy, AbLang, IgLM, PALM, EATLM} pre-trained solely on antibody sequences. Recently, an increasing number of PLMs~\cite{STEPS, PromptProtein, ESM-GearNet, ESM-GearNet-INR, ProstT5, SaProt} have explored the incorporation of protein structures for pre-training, further enriching the pre-training data and obtaining powerful and generalized protein representations. Motivated by their promising endeavors and the significance of structural information, we propose to integrate antibody sequences and structures for multi-level pre-training. \ourapproach pioneeringly makes full use of sequences and structures across multiple domains within a hierarchical pre-training paradigm, significantly contributing to comprehensive antibody representation learning.
Additionally, we customize two objectives to 
% explore the impact of incorporating biological multi-level information mechanisms 
dig out the profound interconnections among biological multi-level information in antibody pre-training. Furthermore, \ourapproach is comprehensively evaluated in extensive experiments across multiple aspects of antibody applications.

\subsection*{Note S5. Ablation Study}
% Extensive ablation experiments from multiple aspects are conducted to evaluate the effectiveness of the proposed hierarchical pre-training paradigm and designed antibody specific pre-training objectives, providing more insights into our approach. 
% % Innovation1
% \subsubsection{Ablation Study on Hierarchical Pre-training paradigm}
% \ourapproach exploits a hierarchical pre-training paradigm, with stage I focused on general sequence-structure learning and stage II dedicated to antibody specific multi-level learning. We first execute the ablation study to evaluate the effect of these two pre-training stages.
% % Innovation2
% \subsubsection{Ablation Study on Multi-level Pre-training Objectives}
% % analysis
% \ourapproach without MLM means pre-training on pairs of antibody sequences and structures (\textit{i.e.} stage II only), and performs poorly among all the ablation experiments. The likely reason is that the limited-scale antibody sequence-structure paired data hampers the model's generalization capability, implying the necessity and significance of the additional protein data in stage I.
% The final three rows of Table~IV present ablation results of the proposed objectives in stage II. 
To evaluate effectiveness of the proposed pre-training objectives, which are tailored for integrating the information from both antibody 1D sequences and 3D structures, we conduct a holistic ablation analysis on multi-type antibody related tasks. The ablation results are reported in Table~\ref{Tab:ablation}. Overall, the performance will decay if any one of the constructed objectives is absent, indicating that both SSM and CLR are essential and advantageous to learn comprehensive antibody representations. In particular, for antibody paratope prediction task, the absence of SSM leads to a more significant performance drop compared to the lack of CLR. This is likely because the matching prediction of 1D and 3Di sequences efficiently injects information from antibody 3D structures that directly determines the spatial binding specificity with antigens. Furthermore, we notice that removing both SSM and CLR for pre-training yields the worst performance among all the ablation experiments. Such phenomenon confirms the superiority of incorporating structural information during the pre-training process and provides more insights into our method.

\begin{table*}[ht]
    \caption{Ablation results of \ourapproach on multi-type antibody related tasks. We primarily validate the effectiveness of designed objectives during \ourapproach pre-training. \textbf{Bold} indicates the best results. The performance will decay if any one of the constructed pre-training objectives is absent, demonstrating that all the objectives provide significant benefits.}
    \centering
    \setlength{\tabcolsep}{2.2em}
    \resizebox{1.0\textwidth}{!}{
    \Large
    \begin{tabular}{c|c c c|c|c c c}
            \toprule
            \multirow{2}{*}{\textbf{}} & \multicolumn{3}{c|}{\textbf{Antigen Binding Prediction}} & \textbf{Maturation Analysis} & \multicolumn{3}{c}{\textbf{Antibody Paratope Prediction}}\\
            \cmidrule{2-8}
             & AUC & F1 & MCC & ACC & AUC & F1 & MCC \\
            \midrule
            \makebox[0.15\textwidth]{Full loss} & \textbf{0.931} & \textbf{0.868} & \textbf{0.705} & \textbf{0.588} & \textbf{0.898} & \textbf{0.708} & \textbf{0.583} \\
            \midrule
            \makebox[0.15\textwidth]{w/o SSM} & 0.916 & 0.858 & 0.681 & 0.576 & 0.889 & 0.677 & 0.566 \\
            \makebox[0.15\textwidth]{w/o CLR} & 0.917 & 0.860 & 0.684 & 0.573 & 0.893 & 0.695 & 0.570 \\
            \makebox[0.15\textwidth]{w/o SSM\&CLR} & 0.902 & 0.847 & 0.664 & 0.562 & 0.882 & 0.671 & 0.552 \\
            \bottomrule
    \end{tabular}}
    \label{Tab:ablation}
\end{table*}

\subsection*{Note S6. Related Work}
% \subsection{Antibody Specific Language Models}
Encouraged by the success of PLMs in protein representation learning, series of works seeks to conduct large-scale pre-training for antibody specific representation learning. AntiBERTy~\cite{AntiBERTy} pioneeringly executes the antibody specific pre-training using 558 million antibody sequences. AntiBERTa~\cite{AntiBERTa} proposes an ALM proficient at the paratope prediction task. AbLang~\cite{AbLang} separately trains Ablang-H and Ablang-L using the heavy-chains and light-chains of antibody sequences to restore missing residues of antibody sequence data. IgLM~\cite{IgLM}, pre-trained on 558 million antibody sequences while conditioning on the chain type and species-of-origin, innovatively creates synthetic libraries by re-designing variable-length spans of antibody sequences. AbBERT~\cite{ABGNN} is trained by predicting masked amino acids in Complementarity Determining Regions (CDRs), which is customized for the antigen-specific antibody design task. EATLM~\cite{EATLM} is the first ALM trying to inject antibody evolution information to language models and demonstrates good performance on the ATUE benchmark. To better leverage knowledge from the field of proteins, IgBert and IgT5~\cite{IgT5} are two ALMs performing antibody sequence pre-training based on the initial weights of ProtBert~\cite{ProtTrans} and ProtT5~\cite{ProtTrans}, and they facilitate the antigen-related binding energy prediction. Recently, PALM~\cite{PALM} demonstrates the remarkable capabilities of ALMs in both the accurate prediction of binding affinity and the diverse generation of antibody candidates. Compared to works in the general protein field~\cite{PromptProtein, ESM-GearNet, ESM-GearNet-INR, ProstT5, SaProt}, there has been limited exploration of integrating information of 1D sequences and 3D structures during large-scale pre-training for comprehensive antibody representation learning. 
% hybrid encoder & large-scale pre-training
AntiBERTa2-CSSP~\cite{AntiBERTa2} exploits the contrastive learning technique to align the feature spaces of antibody sequences and structures, but the absence of a hybrid encoder and large-scale pre-training hinders further progress.
In this paper, \ourapproach proposes a step towards integrating 1D sequences and 3D structures for large-scale pre-training, ultimately learning comprehensive antibody representations.

\printbibliography

\end{document}